\documentclass{article}
\usepackage{spconf,amsmath,graphicx}
\usepackage{booktabs}
\usepackage{graphicx}
\usepackage{subfigure}
\usepackage{enumitem}
\usepackage{ragged2e}
\usepackage{float}
\usepackage{color}
\usepackage{soul}
\usepackage{lineno,hyperref}
\usepackage{cite}

\title{SEVGGNet-LSTM: a fused deep learning model for ECG classification}
\name{Tongyue He$^1$, Yiming Chen$^1$, Junxin Chen$^{2}$,\thanks{Corresponding Author: Junxin Chen (junxinchen@ieee.org). 
		This work is funded by the National Natural Science Foundation of China (No. 62171114).}
	Wei Wang$^3$, Yicong Zhou$^4$}
\address{$^1$College of Medicine and Biological Information Engineering, Northeastern University, Shenyang, China.\\
	$^2$School of Software, Dalian University of Technology, Dalian, China.\\
	$^3$School of Intelligent Systems Engineering, Sun Yat-sen University, Shenzhen, China.\\
    $^4$Department of Computer and Information Science, University of Macau, Macau, China.}
\begin{document}
\begin{sloppypar}
%
\maketitle
\begin{abstract}
This paper presents a fused deep learning algorithm for ECG classification.
It takes advantages of the combined convolutional and recurrent neural network for ECG classification, and the weight allocation capability of attention mechanism.
The input ECG signals are firstly segmented and normalized, and then fed into the combined VGG and LSTM network for feature extraction and classification.
An attention mechanism (SE block) is embedded into the core network for increasing the weight of important features.
Two databases from different sources and devices are employed for performance validation, and the results well demonstrate the effectiveness and robustness of the proposed algorithm.
\end{abstract}
\section{Introduction}
\label{sec:intro}
The electrocardiogram (ECG) analysis is an important non-invasive means for diagnosing and evaluating cardiac diseases \cite{MukhopadhyayAn}. 
However, the inherent complexity of arrhythmia often brings difficulties to medical workers in ECG classification, and may leads to mis-diagnosis.
With the popularity of artificial intelligence (AI), developing computer-aided ECG classification is in a high demand.

Deep learning based solution for ECG classification has drawn world-wide concerns in recent years. 
It is able to automatically extract features, and hence get rid of the dependence of manual feature extraction in traditional machine learning methods.
About features extraction, as reported in  \cite{LaiUnderstandingmore}, attention mechanism is able to increase the weight of important features and further promote the classification performance.
This advantage has been widely exploited in the fields of neural machine translation and computer vision \cite{GuoRe-attention, VaswaniAttention}.
It is therefore plausible to infer that adding attention mechanisms to the ECG classification model is likely to achieve performance improvement.
In addition, previous works \cite{oh2018automated} have demonstrated the great potentials of the combined structure of convolutional neural network (CNN) and rerrent neural network (RNN) in the ECG classification problems. 
As a popular CNN, the visual geometry group network (VGGNet) \cite{simonyan2014very} has more nonlinear transformation and enhanced ability to learn features.
On the other hand, long short-term memory neural network (LSTM) is an improved RNN, and it is able to avoid the problems of gradient explosion or gradient disappearance.

This paper proposes the SEVGGNet-LSTM model for ECG classification.
Our proposal takes advantages of the combined structure of convolutional and recurrent neural network for ECG classification, and also makes full use of the attention mechanism's weight allocation capability.
First, ECG data records are split into 10s segments.
After that, the amplitudes of the divided segments are normalized.
The preprocessed ECG signals are then fed into the SEVGGNet-LSTM, which is a sequential combination of VGG and LSTM, with an attention mechanism (squeeze-and-excitation block, SE block) to increases the weight of important features.
Finally, ECG classification is completed by the fully connected layers.
Two databases from different sources and devices are employed for performance validation, and  the experimental results well demonstrate the effectiveness and advantages of the proposed algorithm.
 
Our main contributions are as follows.
1) A fused deep convolutional neural network is constructed for ECG classification.
2) The combined convolutional and recurrent neural network and the weight allocation capability of the attention mechanism are exploited for performance promotion.
3) Two databases with ECG records collected by different devices are employed for performance evaluation, and the results well demonstrate the advantages of our proposal.

\section{The proposed method}
\label{sec:method}
\subsection{Architecture}
\begin{figure}[t]
	\centering
	\includegraphics[width=6cm]{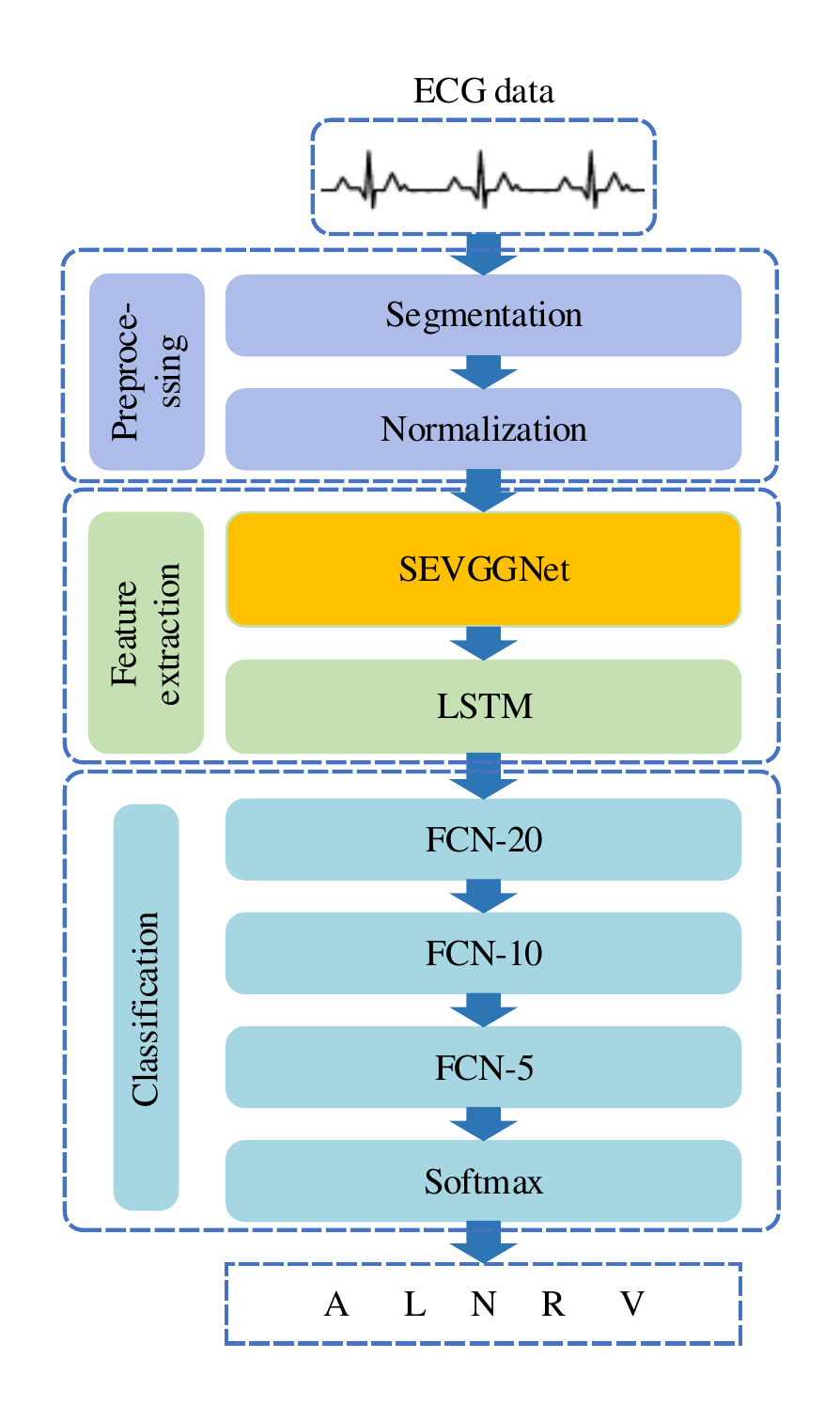}
	\vspace{-0.6cm}
	\caption{\label{fig:algorithm} Block diagram of the proposed model.}
	\vspace{-0.4cm}	
\end{figure}

The proposed algorithm mainly consists of the preprocessing module, feature extraction module, and classification module, as illustrated in Fig.~\ref{fig:algorithm}.
The preprocessing module includes signal segmentation and normalization.
The ECG signals are split into 10s segments, after that a normalization technique is used to normalize their amplitudes.
In the feature extraction module, SEVGGNet and LSTM constitute the deep neural network, and the preprocessed ECG signals are fed into the deep neural network for feature extraction. 
The fully connected layers are finally introduced for classification.

\subsection{Data preprocessing}
The data preprocessing module focus on signal segmentation and normalization.
In order to reduce the computational burden, ECG signals are cut into several 10s segments.
The amplitudes of the ECG segments vary greatly due to the individual differences and lead locations, so they have to be normalized.
The normalization is conducted by
\begin{equation}
	Normalized(X) = \frac{{X - \bar X}}{S},
\end{equation}
where $X$ is  value of each record, $\bar X$ and  $S$ refer to the average and standard deviation of all the records, respectively.

\subsection{Feature extraction}
\begin{figure*}[!ht]
	\centering
	\includegraphics[width=14cm]{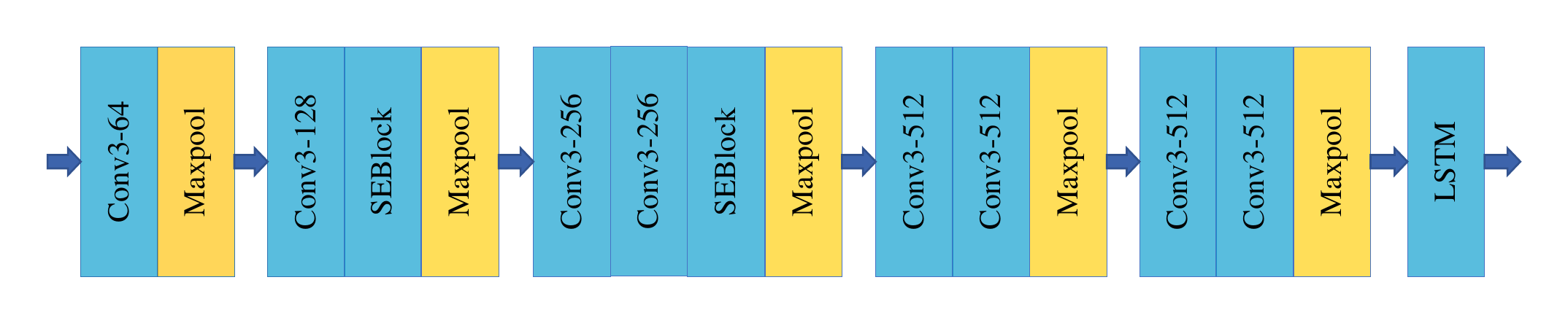}
	\vspace{-0.4cm}
	\caption{\label{fig:corenetwork} Core network of the proposed model.}
	\vspace{-0.4cm}	
\end{figure*}

The core network of the proposed algorithm is illustrated in Fig.~\ref{fig:corenetwork}.
It includes eight convolutional layers, one LSTM layer, five maximum pooling layers, and two SE blocks.
The convolutional layers are split into five parts by maximum pooling layers.
The numbers of the convolutional layers in each part are 1, 1, 2, 2, and 2, and the sizes of convolutional kernels are 64, 128, 256, 512, and 512, respectively.
Each part of convolutional layer is followed by a maximum pooling layer to reduce the data length and computational burden of the model.
A LSTM layer is connected after the convolutional layers and maximum pooling layers of VGGNet to avoid the gradient disappearance and gradient explosion.
In addition, two SE blocks are added to the fusion model of VGGNet and LSTM, which strengthen the features of R peak.
After that, the output of LSTM is given to the fully connected layer for ECG classification.

The SENet, a kind of attention mechanism, is presented in the form of the SE block.
It contains the squeeze and excitation modules.
Squeeze operation is carried out after the traditional convolution module, that is, all spatial features in a channel are encoded into a global feature.
It is defined as 
\begin{equation}
{Z_c} = {F_{sq}}({u_c}) = \frac{1}{{H \times W}}\sum\limits_{i = 1}^H {\sum\limits_{j = 1}^W {{u_c}(i,j),z \in {R^c}} }, 
\end{equation}
where $Z_c$ refers to the $c$-th element of the squeezed channels, $u_c$ represents the $c$-th channel of the input, $F_{sq}$ is the squeeze function, and $H$ and $W$ are the height and width of the input, respectively.
After that, two fully connected layers are employed for better generalization, an activation function is used to obtain the channel-wise dependencies.
The process is described by
\begin{equation}
s = {F_{ex}}(z,W) = \sigma (g(z,W)) = \sigma ({W_2}\delta ({W_1}z)),
\end{equation}
\begin{equation}
\tilde x_c = Fscale({u_c},{s_c}) = {s_c}{u_c},
\end{equation}
where $W_1$ and $W_2$ represent the first and second full-connection operation,  $\delta$ denotes the ReLU activation function, $\sigma$ is the sigmoid function, and $Fscale$ represents channel-wise multiplication, respectively.

\subsection{Classification module}
At the end of our model, a 3-layers fully connected layer of the VGGNet is selected, and the softmax function is used to realize multi-classification problem.
The fully connected layers reassembles the local features into a complete graph through the weight matrix. 
The representative features are integrated into a value, which has the advantage of reducing the influence of feature locations on classification results, and improving the robustness of the whole network.
In addition, multiple fully connected layers are connected, so that the ability of nonlinear expression is improved.

\section{Experiment configuration}
\label{sec:exper}
\subsection{Database and performance metrics}
Two different databases are employed for performance evaluation, the MIT-BIH arrhythmia database and the 2017 PhysioNet/CinC Challeng database (2017PCCD).
The MIT-BIH arrhythmia database includes 48 recordings, each of which is sampled at 360 Hz and contains a series of two-leads ECG data. 
Based on the proportion of various arrhythmias in the clinic, normal rhythm (N), left bundle branch block (R), right bundle branch block (L), ventricular precontraction (V) and atrial premature contractions (A) are selected for classification.
On the other hand, there are 8528 single-lead ECG records in 2017PCCD.
All of them are collected by wearable devices at a sampling frequency of 300 Hz.
The types of normal (N) and atrial fibrillation (AF) samples are selected to validate the performance of our model.
Because of the data imbalance problem, samples in the training set are balanced by oversampling, as detailed in Section \ref{sec:overs}. 

In this paper, accuracy ($Acc$), sensitivity ($Sen$), precision ($Pre$), and F1 score are employed for performance evaluation.
The $Acc$ is the proportion of correctly classified samples, $Sen$ denotes the recognition ability of positive examples, $Pre$ represents the proportion of correct predictions in the samples with positive predictions, and F1 score is the weighted average of $Sen$ and $Pre$, respectively.

All of the models are implemented on the framework Tensorflow-gpu 2.6.2, using the Windows-11 operation system. 
The algorithm is deployed on a workstation with Intel(R) Core(TM) i7-12700H at 2.7 GHz, and an RTX 3060 GPU with a 14 GB memory.

\subsection{Oversampling}
\label{sec:overs}
Oversampling is adopted to solve the data imbalance problem, by increasing the number of certain arrhythmia samples.
In this paper, the random oversampling method is used.
It works as follows. 1) Taking the class with the largest number of samples as the benchmark, calculate the multiple of the benchmark class and the minority classes. 
2) If the multiple is greater than 2, the minority classes are copied by corresponding multiples. 
3) If the multiple is less than 2, a certain proportion (multiple minus one) of samples from the minority classes are randomly selected for duplication.
Taking MIT-BIH arrhythmia database as an example, and samples counts before and after oversampling is listed in  Table~\ref{tab:MITBIH}.

\begin{table}[tb]
	\footnotesize
	\caption{Oversampling on MIT-BIH arrhythmia database}
	\begin{center}
		\setlength{\tabcolsep}{4mm}{
			\begin{tabular}{c c c c}
				\bottomrule
				{Type} & {Before oversampling} & {After oversampling}\\
				\hline
				{N} & {6735} & {6735}\\
				{V} & {3005} & {6010}\\
				{L} & {1202} & {7212}\\
				{R} & {1179} & {7074}\\
				{A} & {771} & {6939}\\
				\hline
				{Total} & {12892} & {33970}\\
				\bottomrule
				\vspace{-1cm}
		\end{tabular}}
		\label{tab:MITBIH}
	\end{center}
\end{table}

\subsection{Cross validation}
Under the condition of limited size of dataset, 10-fold cross validation is able to achieve multiple random partitioning of the training set and test set.
The original dataset is divided into 10 equal sized parts, of which nine parts are considered as training dataset and the other one is used for testing.
In each iteration, the balanced training data set is used to get the optimized parameters of the model, and then the corresponding test set is employed for performance evaluation.
After 10 iterations, the results from each iteration are combined to yield the average performance of the model.

\section{Results and discussion}
\label{sec:resu}
\subsection{Overall performance}
Table \ref{tab:performance} lists the results when using the proposed algorithm to classify various heart rhythms in the MIT-BIH arrhythmia database.
The overall $Acc$, $Sen$, $Pre$ and F1 values are 0.996, 0.984, 0.988, and 0.986, respectively.
By considering both $Sen$ and $Pre$, the F1 score better demonstrate the overall performance of the proposed algorithm.
For A, L, N, R, and V types of rhythms, the achieved F1 scores of our method are 0.955, 0.995, 0.993, 1.000, and 0.987, respectively.
For R rhythm, all of the performance metrics are optimal, that is, 1, which indicates the good performance in identifying R rhythms.
In addition, all of the $Acc$ values are higher than 0.992, demonstrating that the classification performance of our algorithm is very satisfactory.

In order to verify the universality of the proposed algorithm, it is further applied to 2017PCCD, and the performance records are listed in Table \ref{tab:2017results}.
As can be observed, the $Acc$, $Sen$, $Pre$ and overall F1 records are 0.962, 0.931, 0.914, and 0.922, respectively.
As concluded above, the proposed algorithm shows satisfactory performance on both databases.

\begin{table}[tb]
	\footnotesize
	\caption{Performance on MIT-BIH arrhythmia database}
	\begin{center}
		\setlength{\tabcolsep}{2.5mm}{
			\begin{tabular}{c c c c c}
				\bottomrule
				{Types} & {$Acc$} & {$Sen$} & {$Pre$} & {F1} \\
				\hline
				{A} & {0.994} &  {0.955} & {0.955} &  {0.955} \\
				{L} & {0.999} &  {0.991} & {1.000} &  {0.995} \\
				{N} & {0.992} &  {0.997} & {0.989} &  {0.993}  \\
				{R} & {1.000} &  {1.000} & {1.000} &  {1.000} \\
				{V} & {0.994} &  {0.978} & {0.997} &  {0.987} \\
				\hline
				{Overall} & {0.996} &  {0.984} & {0.988} &  {0.986} \\
				\bottomrule
				\vspace{-1cm}
		\end{tabular}}
		\label{tab:performance}
	\end{center}
\end{table}

\begin{table}[t]
	\footnotesize
	\vspace{-0.1cm}
	\caption{Performance on 2017PCCD}
	\begin{center}
		\setlength{\tabcolsep}{2.5mm}{
			\begin{tabular}{c c c c c}
				\bottomrule
				{Types} & {$Acc$} & {$Sen$} & {$Pre$} & {F1} \\
				\hline
				{N} & {0.962} &  {0.974} & {0.982} &  {0.978} \\
				{A} & {0.962} &  {0.888} & {0.845} &  {0.866} \\
				\hline
				{Overall} & {0.962} &  {0.931} & {0.914} &  {0.922} \\
				\bottomrule
				\vspace{-1.2cm}
		\end{tabular}}
		\label{tab:2017results}
	\end{center}
\end{table}

\subsection{Comparison results}
In order to further validate the performance of our proposal, some state-of-the-art algorithms are employed for comparison.
For fair comparison, only the algorithms developed for the same classification problems and tested on the same database are introduced.
Regarding the MIT-BIH arrhythmia database, the selected models for comparison are as follows.
Liu $et~al.$ \cite{liu2022arrhythmia} proposed a model based on LSTM to obtain time-series features of ECG.
In \cite{oh2018automated}, the CNN layer is used to extract feature maps, and the LSTM layer captures the temporal dynamics.
The model \cite{huang2021multiview} is composed of a eight-layers CNN, a eight-layers LSTM and a fully connected layer. 
Detailed performance records of the compared models are listed in Table~\ref{tab:comparisonfive}.
The F1 score, $Acc$, $Sen$ and $Pre$ of the proposed model are higher than those of the compared models.
The advantages of our proposal are therefore validated.

\begin{table}[tb]
	\footnotesize
	\vspace{-0.6cm}
	\caption{Comparison on MIT-BIH arrhythmia database}	
	\begin{center}
		\setlength{\tabcolsep}{2mm}{
			\begin{tabular}{c c c c c }
				\bottomrule
				{Methods} & {$Acc$} & {$Sen$} & {$Pre$} & {F1} \\
				\hline
				{LSTM \cite{liu2022arrhythmia}} & {0.986} &  {0.980} & {0.976} &  {0.978} \\
				{CNN+LSTM \cite{oh2018automated}} & {0.981} & {0.975} & {-} &  {-}\\
				{CNN+LSTM+HOS \cite{huang2021multiview} } & {0.989} &  {0.965} & {0.969} &  {0.967} \\
				{U-Net \cite{oh2019automated}} & {0.973} &  {-} & {-} &  {-}\\
				{RBF-BA \cite{ebrahimzadeh2014detection}} & {0.952} &  {0.956} & {0.906} & {0.930}\\
				{LDA \cite{yeh2009cardiac}} & {0.962} &  {0.925} & {0.947} &  {0.936}\\
				{KNN \cite{jekova2008assessment}} & {-} &  {0.809} & {0.769} &  {0.788}\\
				\hline
				\textbf{{Proposed method}} & \textbf{{0.996}} & \textbf{{0.984}} & \textbf{{0.988}} &  \textbf{{0.986}}\\
				\bottomrule
				\vspace{-1cm}
		\end{tabular}}
		\label{tab:comparisonfive}
	\end{center}
\end{table}

\begin{table}[tb]
	\footnotesize
	\vspace{-0.2cm}
	\caption{Performance comparison on 2017PCCD}
	\begin{center}
		\setlength{\tabcolsep}{2mm}{
			\begin{tabular}{c c c c c}
				\bottomrule
				{Methods} & {$Acc$} & {$Sen$} & {$Pre$} & {F1} \\
				\hline
				{CNN \cite{pourbabaee2018deep}} & {0.899} &  {0.671} & {0.590} &  {0.628} \\
				{VGGNet \cite{simonyan2014very}} & {0.976} & {0.909} & {0.903} & {0.906} \\
				{MS-CNN \cite{fan2018multiscaled}} & {0.977} &  {0.943} & {0.886} &  {0.914} \\
				{BT \cite{mei2018automatic}} & {0.966} &  {0.832} & {-} & {-}\\
				\hline
				\textbf{{Proposed}} & \textbf{{0.962}} &  \textbf{{0.931}} & \textbf{{0.914}} &  \textbf{{0.922}}\\
				\bottomrule
				\vspace{-1cm}
		\end{tabular}}
		\label{tab:comparisontwo}
	\end{center}
\end{table}

\begin{table}[tb]
	\footnotesize
	\vspace{-0.1cm}
	\caption{Classification results on MIT-BIH arrhythmia database}
	\vspace{-0.3cm}
	\begin{center}
		\setlength{\tabcolsep}{2mm}{
			\begin{tabular}{c c c c c c}
				\bottomrule
				{Methods} & {$Acc$} & {$Sen$} & {$Pre$} & {F1}\\
				\hline
				{VGG11-LSTM} & {0.980} &  {0.911} & {0.911} &  {0.911}\\
				{VGG13-LSTM} & {0.980} & {0.894} & {0.919} &  {0.906} \\
				{VGG16-LSTM} & {0.977} &  {0.867} & {0.911} &  {0.888}\\
				{{SEVGG11-LSTM}} & {0.980} &  {0.933} & {0.909} & {0.919}\\
				\hline
				\textbf{{Proposed (SEVGG11-LSTM-O)}} & \textbf{{0.996}} & \textbf{{0.984}} & \textbf{{0.988}} & \textbf{{0.986}}\\
				\bottomrule
				\vspace{-1cm}
		\end{tabular}}
		\label{tab:MITmodels}
	\end{center}
\end{table}

\begin{table}[!h]
	\footnotesize
	\vspace{-0.1cm}
	\caption{Classification results on 2017PCCD}
	\begin{center}
		\setlength{\tabcolsep}{2mm}{
			\begin{tabular}{c c c c c c}
				\bottomrule
				{Methods} & {$Acc$} & {$Sen$} & {$Pre$} & {F1}\\
				\hline
				{VGG11-LSTM} & {0.952} &  {0.841} & {0.950} &  {0.885}\\
				{VGG13-LSTM} & {0.945} & {0.868} & {0.893} &  {0.880} \\
				{VGG16-LSTM} & {0.941} &  {0.798} & {0.950} &  {0.853}\\
				{{SEVGG11-LSTM}} & {0.952} & {0.877} & {0.913} & {0.894}\\
				\hline
				\textbf{{Proposed (SEVGG11-LSTM-O)}} & \textbf{{0.962}} & \textbf{{0.931}} & \textbf{{0.914}} & \textbf{{0.922}}\\
				\bottomrule
				\vspace{-1.2cm}
		\end{tabular}}
		\label{tab:2017models}
	\end{center}
\end{table}

In addition, the two-class classification performance of wearable ECG is also compared on the 2017PCCD.
The results are listed in Table \ref{tab:comparisontwo}. 
Compared with the peer algorithms that have reported their F1 performance, the proposed model has the highest F1 score of 0.922.
The comparison results further verify the good performance of the proposed model.

\subsection{Discussions}
In order to determine the optimized network, convolutional networks of various depths are created.  
In specific, networks with 11, 13 and 16 convolutional layers are compared on the MIT-BIH arrhythmia database.
As listed in Table~\ref{tab:MITmodels}, with the increasing counts of convolutional layers, the performance metrics decrease. 
Compared with the basic models, VGG11-LSTM is finalized for further improvement.
When enhanced by the attention mechanism, the overall F1 score increases, indicating that the SEVGG11-LSTM has better classification performance.
After oversampling is implemented, the classification performance is further improved, and the strong competitiveness on the MIT-BIH arrhythmia database is advantageous to that of the state-of-the-art algorithms.

Similar validation experiments are also performed on the 2017PCCD, and the performance records are comparatively listed in Table~\ref{tab:2017models}. 
Similarly, the F1 score decreases with the increasing counts of the convolutional layers, and is increased after introducing the attention mechanism. 
When oversampling is further used, the F1 score reaches the optimum value.
Tables \ref{tab:MITmodels} and \ref{tab:2017models} well validate the great potentials of attention mechanism and oversampling for ECG classification. 

Compared with the validation on a single database, two databases are employed in our paper in a ``dual-centers" fashion.
More participants are involved to avoid the limitation of a single database, and the conclusion is therefore more reliable. 
In addition, researchers can draw on the wisdom of the masses in the two databases to improve clinical trials.

\section{Conclusion}
\label{sec:conc}
This paper demonstrates a novel deep learning algorithm for ECG classification.
It makes use of the combined convolutional and recurrent neural network for classifying ECG as well as the attention mechanism to assign weights.
The input ECG signals are sequentially segmented and normalized.
After that, the preprocessed signals are fed into the combined VGG and LSTM network for feature extraction and classification.
The core network contains an attention mechanism that increases the weight of significant features. 
Two databases from different sources and devices are employed for performance validation, and the results well demonstrate the effectiveness and advantages of the algorithm.
\bibliographystyle{IEEEbib}
\bibliography{strings,refsICASSP}

\end{sloppypar}
\end{document}